\documentclass{article}

\usepackage{microtype}
\usepackage{graphicx}
\usepackage{subfigure}
\usepackage{booktabs} 
\usepackage{amsmath}		
\usepackage{todonotes}
\usepackage{tabu}
\usepackage{multirow}
\usepackage{wrapfig}

\newcolumntype {+}{ >{\global\let\currentrowstyle\relax }}
\newcolumntype {^}{ >{\currentrowstyle }}
\newcommand {\rowstyle}[1]{\gdef\currentrowstyle{#1} %
	#1\ignorespaces
}

\usepackage{hyperref}

\usepackage[arxiv]{sysml2019}

\sysmltitlerunning{FQ-Conv: Fully Quantized Convolution for Efficient and Accurate Inference}

\graphicspath{{Figures/}}

\begin{document}
\bibliographystyle{plainnat}

\twocolumn[
\sysmltitle{FQ-Conv: Fully Quantized Convolution for Efficient and Accurate Inference}	



\sysmlsetsymbol{equal}{*}

\begin{sysmlauthorlist}
\sysmlauthor{Bram-Ernst Verhoef}{equal,imec}
\sysmlauthor{Nathan Laubeuf}{equal,imec,esat}
\sysmlauthor{Stefan Cosemans}{imec}
\sysmlauthor{Peter Debacker}{imec}
\sysmlauthor{Ioannis Papistas}{imec}
\sysmlauthor{Arindam Mallik}{imec}
\sysmlauthor{Diederik Verkest}{imec}
\end{sysmlauthorlist}

\sysmlaffiliation{imec}{IMEC, Leuven, Belgium}
\sysmlaffiliation{esat}{Departement Electrotechniek (ESAT), KU Leuven, Leuven, Belgium}


\sysmlkeywords{Machine Learning, Quantization}

\vskip 0.3in

\begin{abstract}
Deep neural networks (DNNs) can be made hardware-efficient by reducing the numerical precision of the weights and activations of the network and by improving the network’s resilience to noise. However, this gain in efficiency often comes at the cost of significantly reduced accuracy. In this paper, we present a novel approach to quantizing convolutional neural network. The resulting networks perform all computations in low-precision, without requiring higher-precision BN and nonlinearities, while still being highly accurate. To achieve this result, we employ a novel quantization technique that learns to optimally quantize the weights and activations of the network during training. Additionally, to enhance training convergence we use a new training technique, called gradual quantization. We leverage the nonlinear and normalizing behavior of our quantization function to effectively remove the higher-precision nonlinearities and BN from the network. The resulting convolutional layers are fully quantized to low precision, from input to output, ideal for neural network accelerators on the edge. We demonstrate the potential of this approach on different datasets and networks, showing that ternary-weight CNNs with low-precision in- and outputs perform virtually on par with their full-precision equivalents. Finally, we analyze the influence of noise on the weights, activations and convolution outputs (multiply-accumulate, MAC) and propose a strategy to improve network performance under noisy conditions. 
\end{abstract}

]



\printAffiliationsAndNotice{\sysmlEqualContribution} 

\section{Introduction}

In recent years, there has been a surge of interest in designing accelerator hardware for neural networks. These neural-network accelerators aim to improve the speed and energy efficiency with which the billions of operations in DNNs are performed. The design of NN-accelerators often goes hand in hand with optimizations at the algorithmic level. Such algorithmic optimizations include changing the structure of the network \citep{he2015deep, howard2017mobilenets}, network pruning \cite{lecun1990optimal, han2015deep}, dimensionality reduction of weight matrices \citep{xue2013restructuring}, and combinations thereof. Moreover, each algorithm requires some sort of quantization of its values before being mapped on chip and, if quantized to low precision, this can produce several additional hardware benefits. For instance, low-precision quantization significantly reduces the memory footprint of the network, thereby reducing the on-chip memory and memory transfers needed. Furthermore, extreme quantization can simplify computations considerably: e.g. a network with ternary weights (i.e. -1, 0 or 1) involves only additions, no computationally expensive multiplications. However, very low-precision quantized networks often imply a reduction in accuracy. Also, low-precision DNNs still entail some higher-precision computations, like Batch Normalization(BN) and nonlinear activation functions, which require changing the numerical formats between hardware operations and extra silicon area, and energy for computation. Hence, quantization techniques that maintain good accuracy are required.
Most of today’s DNN accelerators operate in the digital domain \citep{jouppi2017datacenter, moons201714}, but recently there has been an increase in the number of analog designs \citep{ambrogio2018equivalent, guo2017temperature}. Some of the analog accelerators promise to mitigate the classical von Neumann bottleneck by performing most of the computations in the memory. For example, one type of analog hardware implementation uses Ohm’s law to perform multiplications in the memory elements of a crossbar array. Here, the weights are encoded as conductances, but only a limited number of conductances can be stored in each memory device. Following multiplication, the charges are accumulated on the summation line using Kirchhoff’s current law, equivalent to summation of the weighted activations in dot products. Although promising, such analog-compute-in-memory also poses several challenges. This is because the devices that store the weights (memory cells), generate the input activations (e.g. DAC) and convert the analog summed signal back to the digital domain (ADC) are often noisy and require low-precision quantization to be usable or efficient. For these analog accelerators to work, it is therefore crucial that neural networks perform accurately under noisy low-precision conditions. \\ 
This paper makes the following contributions: (1) We propose an effective quantization technique that learns to optimally quantize the weights and activations of the network during training. (2) We train the network to low precision using a new training technique, called gradual quantization. (3) We show that our proposed quantization technique compares favorably to existing techniques. (4) We present a method to remove the higher-precision nonlinearities and BN from the network. (5) We demonstrate the potential of this approach on two additional datasets, the Google speech commands dataset and on CIFAR-100, showing that ternary-weight (2-bit) CNNs with low-precision in- and outputs and no higher-precision BN and nonlinearity perform comparably to their full-precision equivalents with BN. (6) We show that these networks can handle moderate amounts of noise on the weights, activations and outputs of the convolution, making them suitable for analog accelerators.

\section{Related Work}
\paragraph{Learned quantization.} In recent years, a wide range of quantization methods have been proposed, up to methods to ternarize the network weights \citep{li2016ternary, zhu2016trained} or binarize the weights and activations of DNNs \citep{courbariaux2016binarized, rastegari2016xnor, hubara2017quantized}.  Several of these methods utilized the statistical distributions of the weights and activations to propose good quantization methods \citep{li2016ternary, zhu2016trained, cai2017deep}. This statistical approach is a sound approach from an information-theoretic point of view and often works well, but may be sub-optimal from a DNN perspective. For instance, the statistical-quantization approach may not result in the best quantized solution for the network as a whole. This is because the quantization often happens after the network has been trained in full precision, without querying the quantized network if it would choose these or other quantized values if it were allowed to. Furthermore, the distributional assumptions may not be fully accurate and may change for different datasets, across network layers and during training, thereby complicating the statistical approach. Consequently, recent studies have proposed to learn the optimal quantization during training \citep{zhang2018lq, jung2018learning}. Our proposed method is most similar to the recently proposed PACT method for quantizing the outputs of ReLUs \citep{Choi2019PACT}. In that paper, the authors present a way to parametrically learn the clipping range of the ReLU function for optimal quantization. Our method differs from theirs in that our proposed learned quantization method does not have zero gradients for values in the clipping range and can be applied at any position inside the network, which includes the ReLUs, but also for quantizing the weights, quantizing the immediate linear outputs of convolutions and even for quantizing the inputs (e.g. images) into the DNN. In our experiments presented below we will demonstrate the generality of our quantization method.  

\paragraph{Gradual quantization.}Quantizing DNNs to low precision puts strong constraints on the network, constraints that the network often finds hard to adjust to, as evidenced by decreased accuracy. Previous work has tried to ease the transition to low-precision by quantizing different parts of the network in different stages of the training, e.g. by quantizing the initial layers of a DNN before the later layers \citep{baskin2018nice} or by quantizing different parts within layers at different stages \citep{xu2018deep}. The justification for this process is to give the remaining full-precision parts of the network the chance to compensate for the quantization in other network parts. In contrast to these methods, our proposed gradual quantization method quantizes the entire network at once, but gradually lowers the bitwidth of the weights and activations inside the network. Our method is motivated by the observation that it is relatively easy to quantize at high bitwidths and that networks with lower precision can likely learn from networks with slightly higher precision. 

\paragraph{Noise resilience of neural networks.} DNNs have a special relationship with noise. For example, both dropout \citep{srivastava2014dropout} and batch normalization \citep{ioffe2015batch} add noise to the weights or activations during training, thereby improving generalization performance of the network. Importantly, both techniques deactivate the noise source during inference, which is different from what happens in analog accelerators where noise is inherent to the circuitry and thus also present during inference. Other studies have examined the effect of noise on the network weights \citep{merolla2016deep}, finding that (weight-) quantized models better withstand weight noise during inference compared to their full-precision counterparts. Here, we examine the influence of noise on the weights, activations and convolution outputs (MACs) and propose a technique to improve network performance under noisy conditions.

\section{The Proposed Approach}
\paragraph{Overview.} We aim to train convolutional layers of a CNN in which inputs, weights and outputs are fully quantized to low-precision numerical values, in which no higher-precision BN and nonlinearities need to be computed, and for which the resulting CNN performs at high accuracy. To achieve these objectives, we combine three methods during network training: 1. We propose a novel learned quantization technique, 2. We present a new training technique that improves the accuracy of quantized networks, 3. We combine the previous methods with network distillation for best accuracy. In the following sections we will present each method in detail and finally discuss how we eliminate the higher-precision BN and nonlinearities from the network.

\subsection{Learned quantization.} \label{sec:learnQ}
We seek to quantize the inputs, outputs and weights of a convolutional layer in an optimal way using a quantization method that does not rely on any distributional assumptions, can be used for all elements of the network (weights and activations) and gives the network the chance to learn the quantization that is optimal for the entire network, i.e. end-to-end. We will now discuss this in detail.\\
Uniform quantization requires a range in which the values are quantized. Crucially, we do not know if the network relies most on extreme values (i.e. in the tails of the distribution) or small values. Also, the optimal quantization range may change across layers and may be different for weights and activations. We therefore instruct the network to learn the quantization range during training. To do so, we introduce a learnable scale factor in the quantization process. This scale factor can differ per layer and for weights and activations. Our method can be summarized by the following two equations. First, we employ a uniform quantization rule: 
\begin{equation}
quantize \left( x \right)= round\left( clip\left( x, b, 1 \right)\times n\right)/n
\label{q_rule}
\end{equation}

Here, \(x\) can be weights or activations; \(b\) is a lower bound, equal to \(-1\) for quantizing weights/linear outputs of convolutions/inputs to CNNs, and equal to zero for quantized ReLUs (see below); and \(n\) is the number of positive quantization levels, which is \(2^{\left(nb-1\right)}-1\), for \(nb\) bits bitwidth. Thus, we force the quantization to happen in the \([b, 1]\)-range. The quantization scale is then parametrized as follows:
\begin{equation}
Q\left(x\right)=\ e^s\times quantize(\frac{x}{e^s}), 
\label{q_scale}
\end{equation}

where \(s\) is a learnable scale parameter. So, the learnable scale parameter first scales the non-quantized values so that they can be clipped in the standardized \([b,1]\)-range of the quantization function \eqref{q_rule}, after which the quantized result is scaled back to its original range. Therefore, the network can learn whichever range is most optimal for quantization. Note that we employ \(e^s\), i.e. the exponential of \(s\). This function is differentiable in  \(s\) and forces the scaling to be positive. Positive scaling is preferred, otherwise the scaling can, in addition to the network weights, change the sign of the weights and activations, thereby causing training instabilities. Furthermore, positive scaling avoids division by zero.\\ 
The quantization function involves a non-differentiable rounding function, which causes problems when learning the scaling parameter \(s\) with backpropagation. To mitigate this issue, we employ the straight-through-estimator (STE) approach, as introduced by \citep{courbariaux2016binarized, hinton2012neural}. The STE passes the gradient through the non-differentiable rounding function, basically ignoring it in the backward pass. In agreement with \citep{courbariaux2016binarized} we also keep a copy of the non-quantized weights during training and update these non-quantized weights based on the gradient of the quantized weights. The final quantized weights are obtained by quantizing the copy of non-quantized weights.
During the experiments described below, we will use this quantization procedure, allowing each convolutional layer to learn to optimally quantize its weights and activations. 

\begin{figure*}[!h]l
\centering
\includegraphics[width=0.9\linewidth]{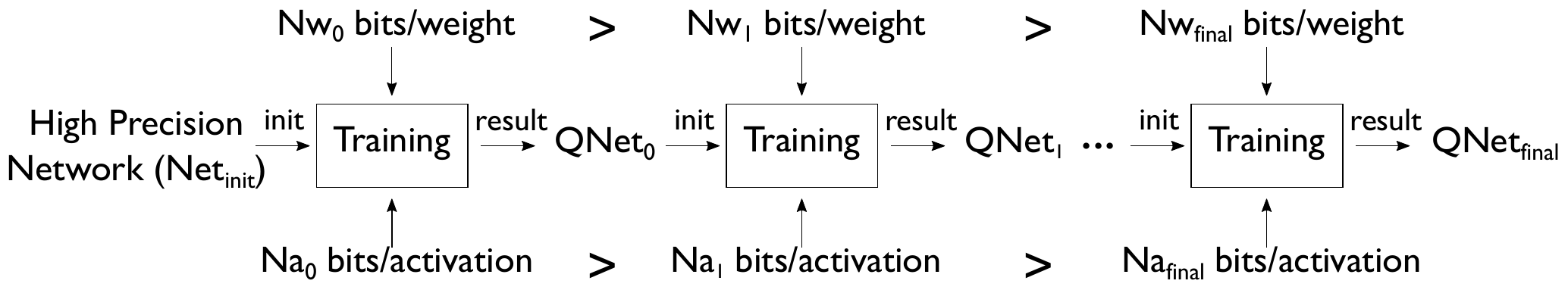}
\caption{Gradual quantization procedure}
\label{fig:gradQuant}
\end{figure*}

\subsection{Gradual quantization}

It is well-known that poor hyper-parameter initialization can cause the network to learn slowly and converge to sub-optimal solutions, exhibiting low accuracy. We found this especially true for networks that are quantized at different positions (quantizing inputs, weights and outputs) and to very low precision (e.g. ternary weights). This is likely because the quantization function \eqref{q_rule} is essentially a saturating nonlinearity, and because a too wide or too narrow initial quantization range effectively collapses all values onto a single quantized value. Both factors increase the likelihood of small gradients during training, and thus poor network convergence. To lessen these issues, and hence improve network convergence, we found it beneficial to gradually lower the bitwidth of the quantization. The general procedure is illustrated in Figure \ref{fig:gradQuant}. Specifically, we start by training a full-precision network and use that network’s trained parameters to initialize a network that is subsequently trained with lower bitwidth (e.g. 8 bits) weights and activations. We then use the parameters of this network to initialize another network with even lower bitwidth (e.g. 5 bits). We continue this procedure until the desired low-bitwidth network is obtained. \\
Gradual quantization is akin to curriculum learning \citep{bengio2009curriculum}, which has been shown to induce better network convergence and improved generalization behavior. Gradual quantization facilitates training, likely because the initialization with networks with similar bitwidths primes the network under training with effective quantization ranges, not too wide or too narrow, such that gradients are larger and learning can happen effectively.\\
Training with gradual quantization takes longer than training once from random initialization, but as soon as the network has been quantized to low precision at high accuracy, it can be deployed and used indefinitely for inference purposes without any additional costs. Finally, remark that similar gradual strategies can help to simplify networks in other ways, besides decreasing the bitwidth (see \ref{sec:noBNReLu}).  

\subsection{Network distillation}
Gradual quantization is a form of transfer learning: it learns from the higher-bitwidth network how to deal with low bitwidths. To further improve the network accuracy, we also include another form of transfer learning, called network distillation. Network distillation was introduced by \citep{hinton2015distilling} to train smaller networks based on larger networks and has subsequently also been used to improve quantized networks \citep{baskin2018nice, polino2018model, wu2016binarized, leroux2019training}. We use the same methods as in \citep{hinton2015distilling}. In short, this technique uses a teacher network to train a student, in our case the low-precision quantized network, by supplying the student network with soft labels, i.e. the output probabilities of the teacher network. The soft labels contain more generalization information than the one-hot training labels (e.g. that a salmon and a goldfish are alike in the sense that they are both fish), which improves test accuracy. This is especially useful for datasets like CIFAR-100, which consist of classes that are subdivided into smaller subclasses. Note that the teacher network does not have to be the full-precision network.

\subsection{Removing the higher-precision BN and nonlinearity}\label{sec:noBNReLu}

\par Batch normalization is used to stabilize the first- and second-order statistics of the activations during training, giving the network the chance to focus learning on more interesting higher order statistics, and improves network convergence during training and generalization behavior during testing \citep{ioffe2015batch}. Batch normalization is performed as: \(
x^{BN}=\gamma\frac{(x-\bar{x})}{S_x}+\beta\), where \(\bar{x}\) and \(S_x\) are the mean and standard deviation of the mini-batch respectively, and \(\gamma\) and \(\beta\) are a learned scale and shift parameter. For inference, the \(\bar{x}\) and \(S_x\) are replaced by their corresponding estimates, \(\widetilde{\mu}\) and \(\widetilde{\sigma}\), based on the complete training dataset. For inference, the BN equation can therefore be simplified as:

\begin{equation}
x^{BN}=\gamma\frac{\left(x-\widetilde{\mu}\right)}{\widetilde{\sigma}}+\beta=\gamma^\prime x+\beta, 
\end{equation}

where \(\gamma^\prime=\gamma/\widetilde{\sigma}\) and \(\beta^\prime=\beta-\frac{\gamma\widetilde{\mu}}{\widetilde{\sigma}}\) . In other words, for inference we only require one scale and one shift factor. Given that the learned quantization method described in \ref{sec:learnQ} already has a scale factor, we can absorb the BN scale factor into the quantization scale factor. We further find that the shift factor doesn’t contribute much to overall accuracy if we train the network to adapt to the absence of the shift factor (see below). This shows that it is possible to remove the BN computations for inference purposes.

\par Next, we observe that quantization function \eqref{q_rule} is a nonlinear function: When the clipping lower-bound \(b\) is set to -1, the quantization function approximates a hard-tanh function: \( y = clip\left(x, -1, 1\right) \). On the other hand, when the lower-bound b is set to 0, the quantization function approximates a ReLU function \(y=\max\left(0,x\right)\). This indicates that we can use the quantization function as a nonlinear activation function.

In practice, we have found it necessary to first train the network to low precision with BNs and nonlinearities in place. Then, once low-precision has been obtained, we initialize a new network with these trained parameters and retrain the network after replacing the combinations of BN+ReLU with the learned quantized ReLU (clipping lower-bound \(b\) set to 0), and isolated BNs with the learned quantization function with clipping lower-bound \(b\) is set to -1 (Figure \ref{fig:noBNReLu}, \ref{fig:QResnet}). During retraining, the learned scale parameters \(s\) are allowed to change, so as to compensate for the new network structure.

\par The resulting FQ-conv layers have quantized inputs, convolve with quantized weights and return quantized outputs, which in turn become the inputs into subsequent FQ-conv layers. No higher-precision BN and activation functions need to be computed. We observe further that the high-precision scale parameters \(s\) are only needed during training to allow the network to learn its optimal quantization. During inference, we can perform integer-valued convolutions. This follows from the definition of the quantization function and the linearity of the dot product: 

\begin{equation}
\begin{aligned}
w \cdot a & = \sum_{i}{Q\left(w_i\right)Q\left(a_i\right)} \\
& = \sum_{i}{s^w\frac{w_i^{int}}{n^w}}s^a\frac{a_i^{int}}{n^a} \\
& = \frac{s^ws^a}{n^wn^a}\sum_{i}{w_i^{int}a_i^{int}}, 
\end{aligned}
\end{equation}

where \(w_i\) and \(a_i\) are weights and activations, and \( Q\left(x\right) \) , \(s^x\) and \(n^x\) are defined as in equation \eqref{q_rule} and \eqref{q_scale} (dropping the exponentials for clarity), and \(w_i^{int}\) and \(a_i^{int}\) are (signed) integer-valued weights and activations, i.e. \(round\left(clip\left(x,\ b,\ 1\right)\times n\right)\). Hence, the multiply-accumulates are performed with integer-valued numbers. Note that for ternary-weight convolutions, with \(w_i^{int}\in\left\{-1, 0, 1\right\}\), only additions/subtractions are performed, and no multiplications. Moreover, the remaining scaling factor \(\frac{s^ws^a}{n^wn^a}\) is not needed for active computation as long as the hardware-supported quantization method (e.g. Lookup tables or Analog-to-digital converters) puts the integer-valued sum into the correct integer-valued quantized bin, which becomes the input into the next layer. \\
The only important scale factor for active computation during inference is the one from the output quantization of the final FQ-Conv layer. This scale factor (\(e^s\)) is applied to the output of the final FQ-conv layer to bring the activations back to the scale expected by the global average pooling layer, which is performed in higher precision.

\section{Experiments}

\subsection{Effectiveness of the proposed quantization technique} 

\par We first examined the effectiveness of the proposed quantization technique. For this purpose, we first employed CIFAR-10 with ResNet-20, a configuration often used to benchmark quantization methods. We trained the network using standard hyper-parameters from previous related studies (learning rate: 0.1, weight decay: 5E-4, 200 epochs, batch size: 128, with standard data augmentation). For proper comparison with the existing literature, we did not quantize the first and last convolutional layer in this analysis (although we do so in subsequent analyses) and report results on the validation set. We also quantized the 1x1 convolutions in the residual paths. 

\par We quantized the network to various bitwidths using gradual quantization, from full-precision down to 2-bit ternary networks (Table \ref{tab:ResNet20GradQ}). We observed test accuracies above (at precisions \(>\)3-bit), equal to (3-bit precision), or slightly below (2-bit precision) a full-precision network trained from scratch (FP0).

\par We further compared test accuracy with and without gradual quantization (GQ). Without GQ, we initialized the network with FP0 parameters and used FP0 as teacher, then quantized immediately to a given low precision. We observed that GQ significantly improves the accuracy of the lowest precision 3-bit and especially 2-bit networks (Table \ref{tab:ResNet20GradQ}). It is likely that one can improve the 2-bit network accuracy without GQ with enough hyper-parameter tuning, here we present an alternative, less error-prone, gradual quantization, technique.

\par We next compared our results to the state of the art (Table \ref{tab:ResNet20Comp}). Our quantization method has lowest degradation compared to FP baseline (DoReFa accuracies taken from \citep{li2016ternary}). For 2-bit networks, LQ-net has slightly higher overall accuracy, despite its increased degradation compared to baseline compared to our method. The overall higher accuracy of LQ-net may be caused by 1. its higher FP baseline, 2. the fact that LQ-net quantizes weights per channel (vs. per layer in our method), i.e.  LG-net uses more learned parameters, 3. LQ-net uses non-uniform quantization (vs. uniform quantization in our method). In sum, we observed that our proposed quantized technique performs well at low precision and compares favorably to existing methods.

\begin{table*}[h]
\centering
\begin{tabular}{@{}cccccccc@{}}
\toprule
Network & \#bits / weight & \#bits / act. & Init. net & Trainer net &Test acc. (\%) & Test acc. No GQ (\%) & Diff (\%) \\
\hline \hline
FP0 & 32 (float) & 32 (float) & - & - & 91.6 & - & - \\
\hline
Q88 & 8 & 8 & FP0 & FP0 & 92.6 & - & - \\
\hline
FP1 & 32 (float) & 32 (float) & Q88 & Q88 & 92.3 & - & - \\
\hline
Q66 & 6 & 6 & Q88 & FP1 & 92.6 & 92.5 & 0.1 \\
\hline
Q55 & 5 & 5 & Q66 & FP1 & 92.6 & 92.5 & 0.1 \\
\hline
Q44 & 4 & 4 & Q55 & FP1 & 92.2 & 92.1 & 0.1 \\
\hline
Q33 & 3 & 3 & Q44 & FP1 & 91.6 & 90.8 & 0.8 \\
\hline
Q22 & 2 & 2 & Q33 & FP1 & 89.9 & 10.0 & 79.9 \\
\bottomrule
\end{tabular}
\caption{Gradual Quantization of ResNet-20 on CIFAR-10}
\label{tab:ResNet20GradQ}
\end{table*}

\begin{table*}[h]
\centering
\begin{tabular}{@{}cccc@{}}
\toprule
Name & Baseline (\%) & Quantized (\%) & Diff (\%) \\
\hline \hline
PACT-SAWB (W2/A2) & 91.5 & 89.2 & 2.3 \\
\hline
LQ-Net (W2/A2) & 92.1 & 90.2 & 1.9 \\
\hline
DoReFa (W2/A2) & 91.5 & 88.2 & 3.3 \\
\hline
\rowstyle{\bfseries}
GQ (W2/A2) & 91.6 & 89.9 & 1.7 \\
\hline
LQ-Net (W3/A3) & 92.1 & 91.6 & 0.5 \\
\hline
\rowstyle{\bfseries}
GQ (W3/A3) & 91.6 & 91.6 & 0.0 \\
\bottomrule
\end{tabular}
\caption{CIFAR-10: Comparison of validation accuracy for ResNet-20. (W/A) gives \# bits for weights/activations.}
\label{tab:ResNet20Comp}
\end{table*}

\par Thus far, we have examined the effectiveness of the proposed quantization technique on a relatively simple problem. We further extended the examination by quantizing DarkNet-19 \citep{redmon2016yolo9000} on ImageNet/ILSVRC2012 \citep{deng2014imagenet}. We used the same training methods as described in \citep{redmon2016yolo9000}, but with only random crops and random horizontal flips as data augmentation. During quantization we used a trained full-precision ResNet-50 as the teacher and applied label refinery \citep{bagherinezhad2018label} with it. Label refinery is similar to network distillation but avoids tuning a temperature hyper-parameter. Like before, the first and last layer were not quantized to low precision but left in full precision. In all other layers the weights and activations were quantized to low precision. Models were trained on eight V100 GPUs using distributed data parallel and the V100 tensor cores (mixed precision training). 

We show the top-1 and top-5 accuracy for quantized models at different low precisions in Table \ref{tab:DarkNet19}. Due to the teacher model and very little effect of low-precision quantization on the validation accuracy, all models except for the ternary-weight model (Q25) achieve better accuracy compared to the full precision model trained from scratch. Even for the ternary-weight model we observe only a moderate (2.4\%/1.3\%) drop in accuracy. 

\begin{table*}[!ht]
\centering
\begin{tabular}{@{}ccccccc@{}}
\toprule
Network & \#bits / weight & \#bits / act. & Init. net & Top-1 (\%) & Top-5 (\%) & Diff (\%) \\
\hline \hline
FP0 & 32 (float) & 32 (float) & - & 72.3 & 90.7 & 0.0/0.0 \\
\hline
Q88 & 8 & 8 & FP0 & 73.7 & 91.6 & -1.4/-0.9 \\
\hline
Q77 & 7 & 7 & Q88 & 73.8 & 91.7 & -1.5/-1.0 \\
\hline
Q66 & 6 & 6 & Q77 & 73.8 & 91.6 & -1.5/-0.9 \\
\hline
Q55 & 5 & 5 & Q66 & 73.4 & 91.4 & -1.1/-0.7 \\
\hline
Q45 & 4 & 5 & Q55 & 73.0 & 91.3 & -0.7/-0.6 \\
\hline
Q35 & 3 & 5 & Q45 & 72.6 & 90.9 & -0.3/-0.2 \\
\hline
Q25 & 2 & 5 & Q35 & 69.9 & 89.4 & 2.4/1.3 \\
\bottomrule
\end{tabular}
\caption{Quantized DarkNet-19 on ImageNet}
\label{tab:DarkNet19}
\end{table*}

\subsection{Keyword spotting with the Google speech commands dataset}

\par We first evaluate FQ-Conv layers on the Google speech commands dataset \citep{warden2017speech}, a typical benchmark dataset for edge applications. The dataset consists of 65K audio clips, each 1sec long, of 30 keywords uttered by thousands of different people. The goal is to classify each audio clip into one of 10 keyword categories (e.g. “Yes”, “No”, “Left”, “Right”, etc.), or a “silence” (i.e. no spoken word, but background noise is possible) or “unknown” (i.e. a class consisting of the remaining 20 keywords from the dataset) category. The dataset was split into 80\% training, 10\% validation and 10\% test data, based on the SHA1-hashed name of the audio clips. Following Google’s preprocessing procedure, we add background noise to each training sample with a probability of 0.8, where the type of noise is randomly sampled from the background noises provided in the dataset. We also add random time shifts \(\sim Uniform(-100, 100)\). From the augmented audio samples, 39-dimensional Mel-Frequency Cepstrum Coefficients (MFCCs; 13 MFCCs and their first- and second-order deltas) are then constructed using 20ms sliding window, shifted by 10ms. These spectral features provide the inputs into the network. 

\begin{figure*}[t]
\centering
\includegraphics[width=0.9\linewidth]{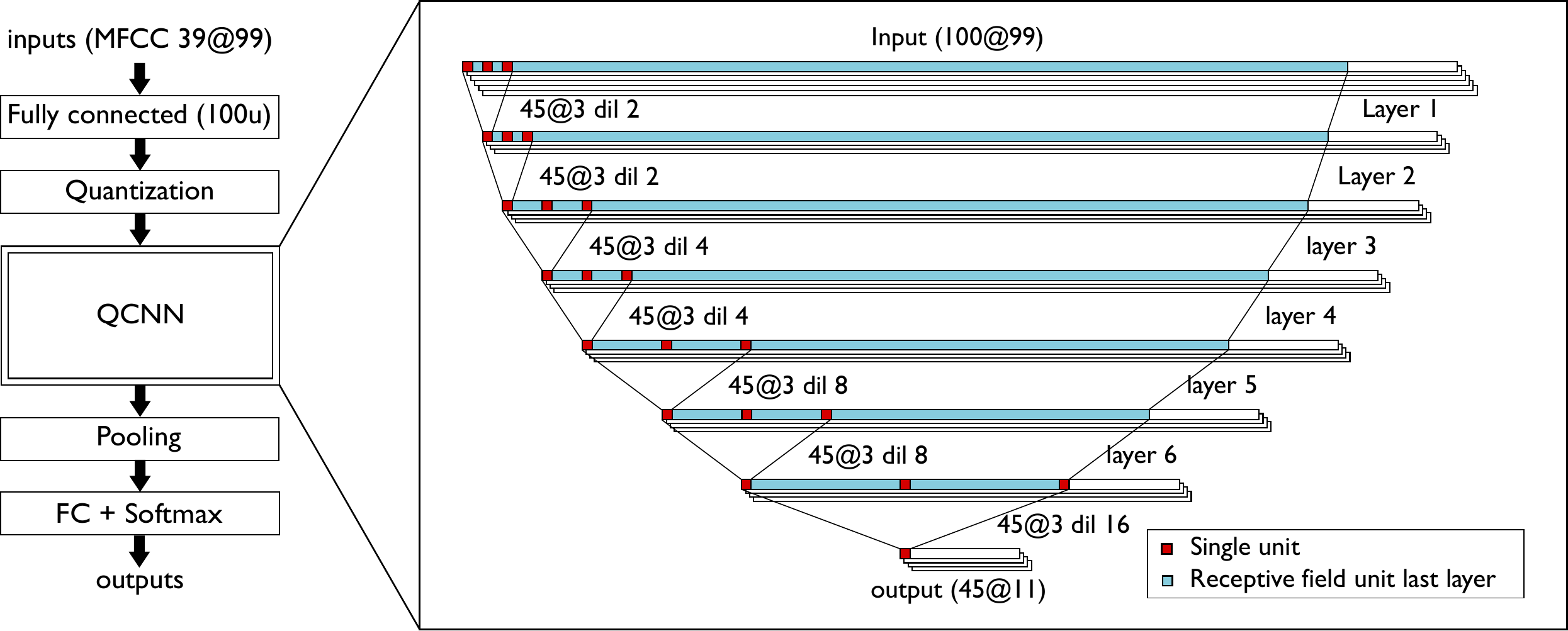}
\caption{Keyword spotting network architecture.}
\label{fig:keyspotnet}
\end{figure*}

\par The network for this application is illustrated in Figure \ref{fig:keyspotnet}. It was developed to have low computational and memory complexity, while still being accurate. The MFCC components are first fed into a small full-precision fully-connected layer (N=100 units). This small (3.9K weights/MACs) layer serves as an expansive embedding of the spectral features such that no input-feature information is lost after quantizing this layer’s output. The output of this layer is batch normalized and quantized to 4 bits before entering the quantized CNN (QCNN) with 7 FQ-Conv layers. Each FQ-Conv layer is a 1D-convolutional layer (45 filters, filter length=3), with no zero-padding  applied. To widen the receptive fields of the units in the final FQ-Conv layer, we employ dilated filters with an exponential-sizing dilation across layers, as shown in Figure \ref{fig:keyspotnet}. The output of the QCNN is global-average pooled before entering the final softmax layer. The network contains 50K parameters and computes 3.5M MACs per sample. 

\par The network was implemented in Pytorch and trained with the ADAM optimizer on Nvidia Tesla V100 GPUs for 600 epochs (batch size of 100). For the full-precision network, the initial learning rate was set to 0.01 and exponentially decayed (decay factor=0.98; network randomly initialized). The network with best performance on the validation set was retained (94.3\% on test set). The full-precision (FP) network served as the initial teacher network and as initialization for the gradual quantization. Each time we obtained a more accurate network on the validation dataset, the more accurate network became the teacher for subsequent networks.

\par For gradual quantization, we used the quantization sequence presented in Table \ref{tab:QkeyTraining}. The table shows the accuracy on the test dataset for each step in the gradual-quantization sequence, where each step is defined by the number of bits used for the weights and activations. The final quantized network has ternary weights and 4-bit activations and a 94.26\% accuracy on the test set, on par with the full-precision network. 

\begin{figure}[!h]
\centering
\includegraphics[width=\linewidth]{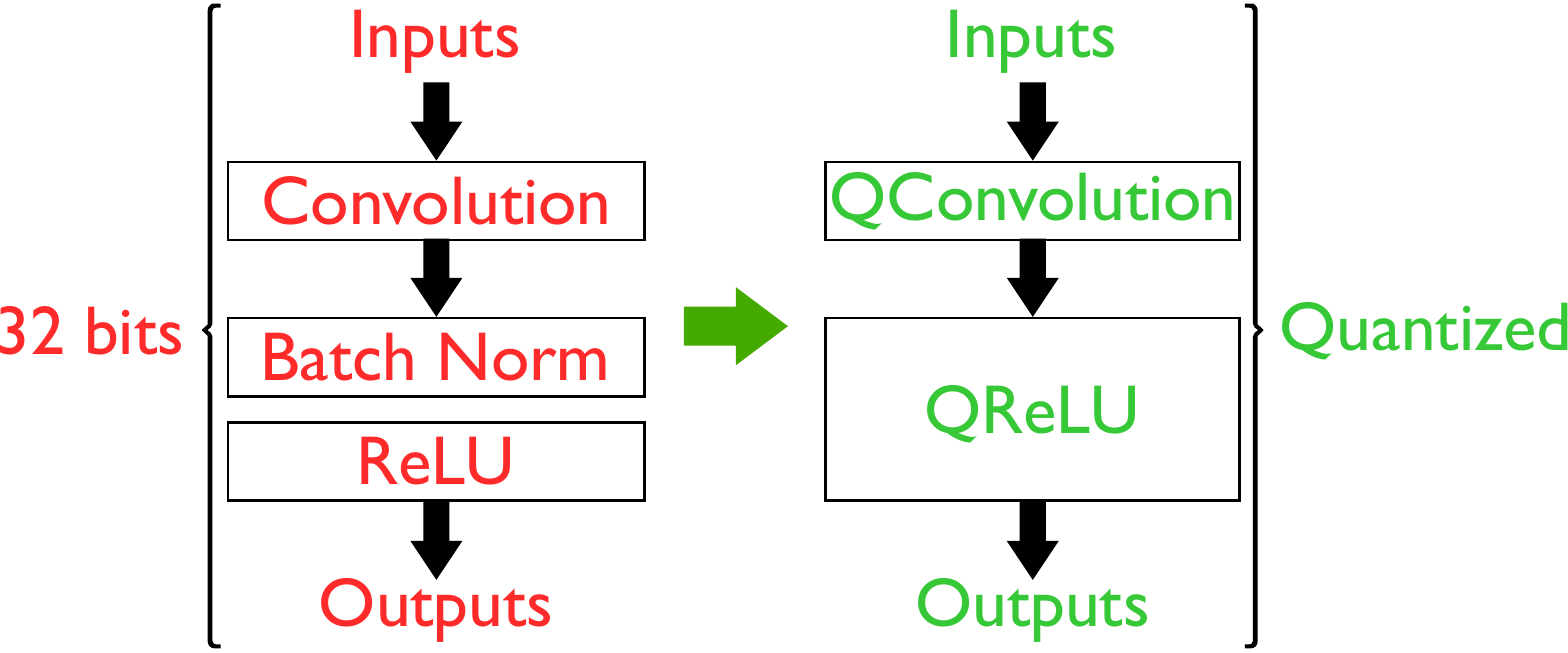}
\caption{Replacing BN+ReLU by a learned quantized ReLU}
\label{fig:noBNReLu}
\end{figure}

\begin{table*}
\centering
\begin{tabular}{@{}cccccc@{}}
\toprule
Network & \#bits/weight & \#bits/activ. & \vspace{2pt} Initializing network \vspace{2pt} & Trainer network & Test accuracy (\%) \\
\hline \hline
FP & 32 (float) & 32 (float) & - & - & 94.3 \\
\hline
Q66 & 6 & 6 & FP & FP & 94.42 \\
\hline
Q45 & 4 & 5 & Q66 & Q66 & 94.68 \\
\hline
Q35 & 3 & 5 & Q45 & Q45 & 94.97 \\
\hline
Q24 & 2 & 4 & Q35 & Q45 & 94.26 \\
\hline
FQ24 & 2 & 4 & Q24 & Q45 & 93.81 \\
\bottomrule
\end{tabular}
\caption{Quantized keyword spotting network training sequence}
\label{tab:QkeyTraining}
\end{table*}

\par In the previous networks, each quantized convolution was followed by a BN+ReLU. Next, we replaced the BN+ReLUs with Quantized ReLUs (\ref{sec:learnQ}) (Figure \ref{fig:noBNReLu}), turning it into a fully quantized Conv-layer. To do so, we initialized the network consisting of FQ-Conv layers with the final parameters obtained with gradual quantization and finetuned (learning rate=0.0005; decay=0.98; 600 epochs) the network. The final network with the fully quantized CNN has an accuracy on the test dataset of 93.81\% (Table \ref{tab:QkeyTraining}), almost as good as the full-precision network with BN, and outperforming several of the larger and higher bitwidth models in the literature  \citep{zhang2017hello, tang2018deep}. 

\begin{table*}
\centering
\begin{tabular}{@{}ccccc@{}}
\toprule
Model & \vspace{2pt}Test accuracy (\%)\vspace{2pt} & \# params & Size (Byte) & Mult. \\
\hline \hline
trad-fpool13 & 90.5 & 1.37M & 5.48M & 125M \\
tpool2 & 91.7 & 1.09M & 4.36M & 103M \\
one-stride1\vspace{2pt} & 77.9 & 954K & 3.82M & 5.76M \\
\hline
res15 & 95.8 & 238K & 952K & 894M \\
res15-narrow\vspace{2pt} & 94.0 & 42.6K & 170K & 160M \\
\hline
Q35 & 94.97 & 50K & 18.75K & 3.5M \\
FQ24 & 93.81 & 50K & 12.5K & 3.5M \\
\bottomrule
\end{tabular}
\caption{Comparison of different keyword spotting models}
\label{table:compare}
\end{table*}

In Table \ref{table:compare}, we compare our best and final low-precision network to some of the best full-precision models reported in the literature \citep{sainath2015convolutional}, \citep{tang2018deep}. Our models have a much smaller memory footprint, require significantly less operations and perform very competitively accuracy-wise.  

\subsection{Visual object classification with CIFAR-100}

\begin{figure*}[h]
\centering
\includegraphics[width=0.8\linewidth]{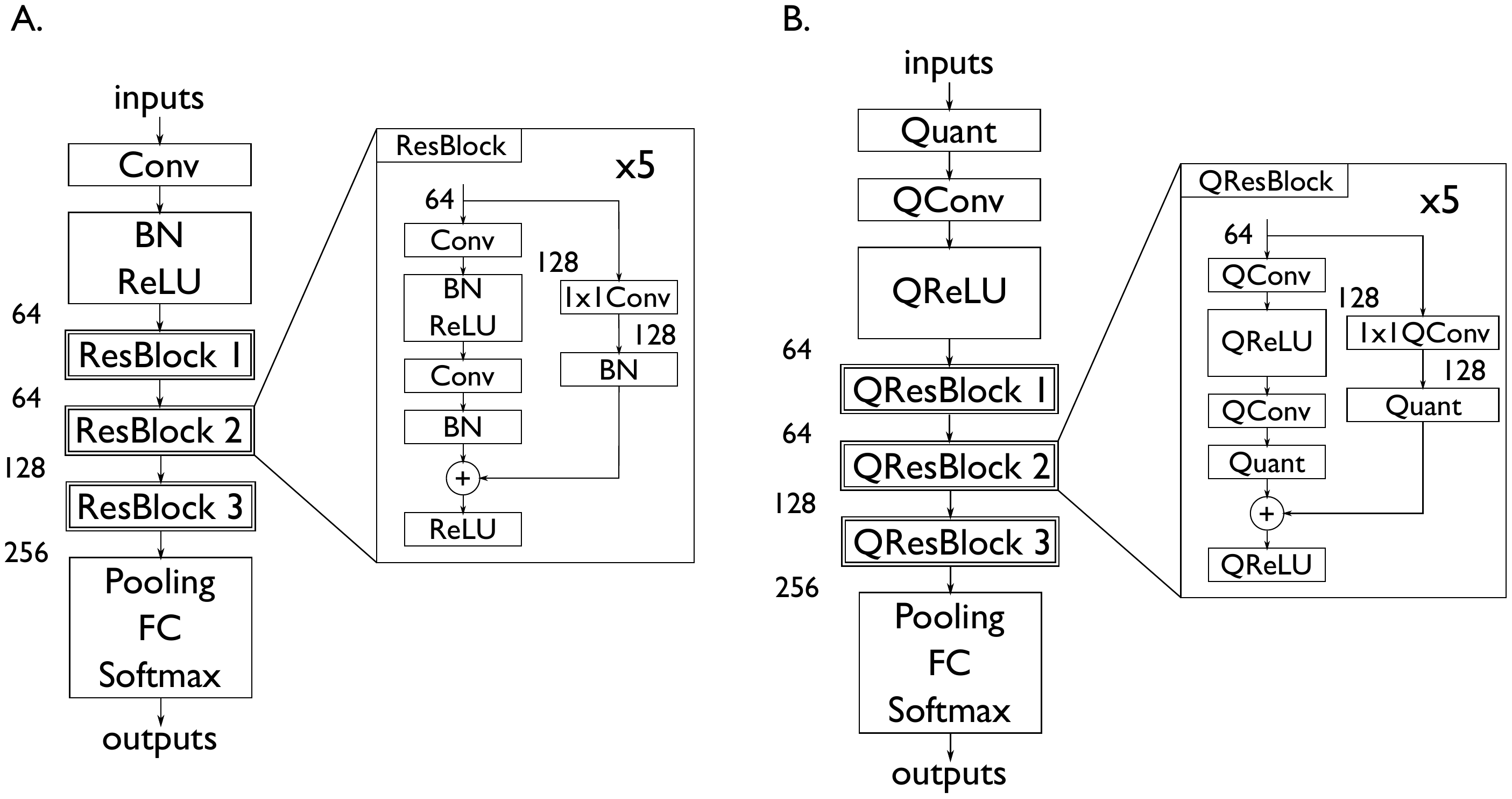}
\caption{Quantized ResNet architecture: A. General architecture, B. Fully quantized architecture. }
\label{fig:QResnet}
\end{figure*}

\par We conducted further studies on the CIFAR-100 dataset, using a ResNet-32 network. The CIFAR-100 dataset comprises 50K training and 10K testing 32 x 32 RGB images in 100 classes. All images were normalized to zero mean and unit standard deviation. For data augmentation, we performed random horizontal flips and random crops from images zero-padded with 4 pixels on each side.

\par The ResNet-32 architecture is shown in Figure \ref{fig:QResnet}A. It consists of a first convolutional layer, followed by BN and ReLU. The output of this layer is fed into three ResBlocks with increasing numbers of filters (64 to 256). Each ResBlock consists of five subblocks with standard residual architecture \citep{he2015deep}, using 1x1 convolution + BN for down sampling between ResBlocks. When quantizing the network, all convolutional layers were quantized (the pooling and softmax layer were left in full precision). The overall architecture of the fully quantized ResNet-32 is shown in Figure \ref{fig:QResnet}B. Note that we also quantized the first conv layer and the 1x1 convolutions in the residual connections. Moreover, the input images are also quantized to lower precision, using learned quantization, before entering the first quantized conv-layer. 

\par The network was implemented in Pytorch and trained with SGD with Nesterov Momentum (0.9 momentum) on V100 GPUs for 200 epochs (batch size of 128), applying a small amount of weight decay (5E-4). We decayed the learning rate by 0.2 after 60, 120 and 180 epochs and report the final test accuracy. For the initial full-precision network, the initial learning rate was set to 0.1, but an initial learning rate of 0.01 was used for gradual quantization and fine-tuning. To obtain a good teacher network, we first trained a full-precision (FP) network from random initialization (top-1: 77.94\%; top-5: 94.43\%), then trained an 8-bit network with the FP-network as initialization and teacher (top-1: 79.82\%; top-5: 94.50\%), and finally trained again an FP-network with the 8-bit network as initialization and teacher (top-1: 79.81\%; top-5: 95.09\%). This final FP-net served as teacher throughout subsequent analyses.

\par For gradual quantization of ResNet-32, we used the quantization sequence presented in Table \ref{tab:QResNetTraining}. The final quantized network has ternary weights and 5-bit activations with a top-1 accuracy of 76.80\% and a top-5 accuracy of 93.53\%. 

\begin{table*}
\centering
\begin{tabular}{@{}ccccccc@{}}
\toprule
Network & \#bits / weight & \#bits / act. & Init. net & Trainer net & top 1 acc.(\%) & top 5 acc. (\%) \\
\hline \hline
FP0 & 32 (float) & 32 (float) & - & - & 77.94 & 94.43 \\
\hline
Q88 & 8 & 8 & FP0 & FP0 & 79.82 & 94.50 \\
\hline
FP1 & 32 (float) & 32 (float) & Q88 & Q88 & 79.81 & 95.09 \\
\hline
Q66 & 6 & 6 & Q88 & FP1 & 78.54 & 94.58 \\
\hline
Q55 & 5 & 5 & Q66 & FP1 & 78.38 & 94.18 \\
\hline
Q45 & 4 & 5 & Q55 & FP1 & 77.96 & 94.26 \\
\hline
Q35 & 3 & 5 & Q45 & FP1 & 77.31 & 93.90 \\
\hline
Q25 & 2 & 5 & Q35 & FP1 & 76.80 & 93.53 \\
\hline
FQ25 & 2 & 5 & Q25 & FP1 & 76.89 & 94.32 \\
\bottomrule
\end{tabular}
\caption{Gradual Quantization of ResNet-32 on CIFAR-100}
\label{tab:QResNetTraining}
\end{table*}

\par  In the previous networks, each quantized convolution was followed by a BN+ReLU or BN (Figure \ref{fig:QResnet}A). To obtain the fully quantized network structure presented in Figure \ref{fig:QResnet}B, we next replaced each BN+ReLU with a Quantized ReLU and the isolated BNs with a learned quantization function with clipping lower-bound b set to -1 (\ref{sec:learnQ}). Subsequently, we initialized the network consisting of FQ-Conv layers with the final parameters obtained from gradual quantization and finetuned the network. The final network with the fully quantized CNN without high-precision BN and ReLUs obtains a top-1 accuracy of 76.89\% and a top-5 accuracy of 94.32\% (Table \ref{tab:QResNetTraining}), close to the FP-network, when trained from random initialization. 

\par It has been shown in previous studies that the first conv-layer (with quantized inputs) and the residual connections are not easily quantized to low-precision without sacrificing too much accuracy \citep{courbariaux2016binarized, rastegari2016xnor, cai2017deep, Choi2019PACT, baskin2018nice, anderson2017high}. Hence it is likely that one can quantize the activations to lower than 5 bits, while retaining high accuracy, if one were to give higher precision to these critical paths compared to the other conv-layers. In this work, we demonstrate the principle and use the extreme case of completely uniform conv-blocks, each with ternary weights and 5-bit activations, and obtain accuracies close to the full-precision network when trained from scratch. Depending on the particular application and hardware, one can adjust the bitwidths in different blocks for optimal performance. 

\subsection{Network performance with additional noise}
\par In a final experiment, we examined the effect of adding additional noise (on top of the quantization noise) to the weights, activations and outcomes of the convolutions (MACs) on the accuracy of the KWS and CIFAR-100 networks. In the context of typical analog accelerators, adding noise to the weights, activations and MACs corresponds to noisy memory cells, DACs and ADCs respectively. Exploring the entire noise space is impossible, so we restrict our exploration to a set of physically plausible noise values, including relatively low and high noise levels. Specifically, we added Gaussian noise (\(\sim \mathcal{N}(0,\,\sigma)\,\)) to the different elements of the network. The amount of noise is quantified by \(\sigma\), which is expressed as a percentage of the least significant bit (LSB). In other words, \(\sigma\) is a percentage of the quantization interval. We used the ternary networks for these experiments.

\par Table \ref{tab:NoiseEffect} presents the network accuracy for different levels of weight noise (\(\sigma_w\)), activation noise (\(\sigma_a\)) and MAC noise (\(\sigma_{MAC}\) for the KWS and CIFAR-100 dataset. We examined network accuracy under different conditions: with or without training with noise. For each condition, we averaged accuracy across ten repetitions (with different noise) of the test set.

\par As expected, small amounts of noise had little influence on network accuracy, but larger amounts of noise clearly lowered test accuracy. However, by training with noise, we could recover much of the accuracy drop (Table \ref{tab:NoiseEffect}).

\begin{table*}
\centering
\begin{tabular}{@{}lllll@{}}
\toprule
Dataset & \multicolumn{2}{c}{KWS} & \multicolumn{2}{c}{CIFAR-100} \\
\hline 
Baseline (No added noise) & \multicolumn{2}{c}{94.3\%} & \multicolumn{2}{c}{76.8\%} \\
\hline 
Test Condition & Not trained with noise  & Trained with noise & Not trained with noise & Trained with noise \\
\hline
\begin{tabular}{@{}l@{}}
\(\sigma_w = 1\%,\) \\ \(\sigma_a = 1\%,\) \\ \( \sigma_{MAC} = 5\%\)
\end{tabular} & 94.3\% & 94.4\% & 76.9\% & 77\%  \\
\hline
\begin{tabular}{@{}l@{}}
\(\sigma_w = 5\%,\) \\ \( \sigma_a = 5\%,\) \\ \( \sigma_{MAC} = 25\%\) \\
\end{tabular} & 94.2\% & 94.6\% & 76.6\% & 76.9\%  \\
\hline
\begin{tabular}{@{}l@{}}
\(\sigma_w = 10\%,\) \\ \( \sigma_a = 10\%,\) \\ \( \sigma_{MAC} = 50\%\)
\end{tabular} & 93.1\% & 94\% & 73.8\% & 75.4\%  \\
\hline
\begin{tabular}{@{}l@{}}
\(\sigma_w = 20\%,\) \\ \( \sigma_a = 20\%,\) \\ \( \sigma_{MAC} = 100\%\)
\end{tabular} & 79.7\% & 91.6\% & 65.1\% & 72.5\%  \\
\hline
\begin{tabular}{@{}l@{}}
\(\sigma_w = 30\%,\) \\ \( \sigma_a = 30\%,\) \\ \( \sigma_{MAC} = 150\%\)
\end{tabular} & 38.8\% & 87.7\% & 34.8\% & 69.2\%  \\
\bottomrule
\end{tabular}
\caption{Effect of noisy weights, activations and MAC results on the accuracy of ternary networks.}
\label{tab:NoiseEffect}
\end{table*}

\section{Conclusion}
This paper presents a novel learned quantization method, a new gradual quantization training strategy and an approach to eliminate high-precision BN and nonlinearities from the network. The result is a network consisting of convolutional layers, in which the weights, inputs and outputs are fully quantized to low precision and high-precision BN and nonlinearities are removed. The accuracy of this low precision network closely approximates that of its full-precision equivalent, which includes BN and higher precision nonlinearities. These low-precision networks are ideal to run in a memory-, computationally-, and energy-efficient way on modern neural-network-accelerator hardware and microcontrollers. Although such low-precision networks can improve the efficiency of both digital and analog accelerator designs, we believe that analog designs especially will benefit from them. For example, in contrast to digital accelerators, the summation of weighted activations in the analog domain has virtually infinite precision and comes at no additional cost, i.e. no higher-precision accumulators are required. Thus, quantizing the inputs, weights and MAC results is sufficient. We further show that these quantized networks tolerate noise quite well. Consequently, these findings suggest that networks implemented on analog arrays can be accurate, fast and efficient. 

\bibliography{references}

\begin{thebibliography}{37}
\providecommand{\natexlab}[1]{#1}
\providecommand{\url}[1]{\texttt{#1}}
\expandafter\ifx\csname urlstyle\endcsname\relax
  \providecommand{\doi}[1]{doi: #1}\else
  \providecommand{\doi}{doi: \begingroup \urlstyle{rm}\Url}\fi

\bibitem[Ambrogio et~al.(2018)Ambrogio, Narayanan, Tsai, Shelby, Boybat, Nolfo,
  Sidler, Giordano, Bodini, Farinha, et~al.]{ambrogio2018equivalent}
Stefano Ambrogio, Pritish Narayanan, Hsinyu Tsai, Robert~M Shelby, Irem Boybat,
  Carmelo Nolfo, Severin Sidler, Massimo Giordano, Martina Bodini, Nathan~CP
  Farinha, et~al.
\newblock Equivalent-accuracy accelerated neural-network training using
  analogue memory.
\newblock \emph{Nature}, 558\penalty0 (7708):\penalty0 60, 2018.

\bibitem[Anderson and Berg(2017)]{anderson2017high}
Alexander~G Anderson and Cory~P Berg.
\newblock The high-dimensional geometry of binary neural networks.
\newblock \emph{arXiv preprint arXiv:1705.07199}, 2017.

\bibitem[Bagherinezhad et~al.(2018)Bagherinezhad, Horton, Rastegari, and
  Farhadi]{bagherinezhad2018label}
Hessam Bagherinezhad, Maxwell Horton, Mohammad Rastegari, and Ali Farhadi.
\newblock Label refinery: Improving imagenet classification through label
  progression.
\newblock \emph{arXiv preprint arXiv:1805.02641}, 2018.

\bibitem[Baskin et~al.(2018)Baskin, Liss, Chai, Zheltonozhskii, Schwartz,
  Girayes, Mendelson, and Bronstein]{baskin2018nice}
Chaim Baskin, Natan Liss, Yoav Chai, Evgenii Zheltonozhskii, Eli Schwartz, Raja
  Girayes, Avi Mendelson, and Alexander~M Bronstein.
\newblock Nice: Noise injection and clamping estimation for neural network
  quantization.
\newblock \emph{arXiv preprint arXiv:1810.00162}, 2018.

\bibitem[Bengio et~al.(2009)Bengio, Louradour, Collobert, and
  Weston]{bengio2009curriculum}
Yoshua Bengio, J{\'e}r{\^o}me Louradour, Ronan Collobert, and Jason Weston.
\newblock Curriculum learning.
\newblock In \emph{Proceedings of the 26th annual international conference on
  machine learning}, pages 41--48. ACM, 2009.

\bibitem[Cai et~al.(2017)Cai, He, Sun, and Vasconcelos]{cai2017deep}
Zhaowei Cai, Xiaodong He, Jian Sun, and Nuno Vasconcelos.
\newblock Deep learning with low precision by half-wave gaussian quantization.
\newblock In \emph{Proceedings of the IEEE Conference on Computer Vision and
  Pattern Recognition}, pages 5918--5926, 2017.

\bibitem[Choi et~al.(2019)Choi, Venkataramani, Srinivasan, Gopalakrishnan,
  Wang, and Chuang]{Choi2019PACT}
Jungwook Choi, Swagath Venkataramani, Vijayalakshmi Srinivasan, Kailash
  Gopalakrishnan, Zhuo Wang, and Pierce Chuang.
\newblock Accurate and efficient 2-bit quantized neural networks.
\newblock Technical report, 2019.
\newblock URL \url{https://www.sysml.cc/doc/2019/168.pdf}.

\bibitem[Courbariaux et~al.(2016)Courbariaux, Hubara, Soudry, El-Yaniv, and
  Bengio]{courbariaux2016binarized}
Matthieu Courbariaux, Itay Hubara, Daniel Soudry, Ran El-Yaniv, and Yoshua
  Bengio.
\newblock Binarized neural networks: Training deep neural networks with weights
  and activations constrained to+ 1 or-1.
\newblock \emph{arXiv preprint arXiv:1602.02830}, 2016.

\bibitem[Deng et~al.(2014)Deng, Su, Krause, Satheesh, Ma, Huang, Karpathy,
  Khosla, Bernstein, Berg, et~al.]{deng2014imagenet}
Jia Deng, Hao Su, Jonathan Krause, Sanjeev Satheesh, Sean Ma, Zhiheng Huang,
  Andrej Karpathy, Aditya Khosla, Michael Bernstein, Alexander~C Berg, et~al.
\newblock Imagenet large scale visual recognition challenge.
\newblock \emph{arXiv preprint arXiv:1409.0575}, 2014.

\bibitem[Guo et~al.(2017)Guo, Bayat, Prezioso, Chen, Nguyen, Do, and
  Strukov]{guo2017temperature}
Xinjie Guo, F~Merrikh Bayat, Mirko Prezioso, Y~Chen, B~Nguyen, N~Do, and
  Dmitri~B Strukov.
\newblock Temperature-insensitive analog vector-by-matrix multiplier based on
  55 nm nor flash memory cells.
\newblock In \emph{2017 IEEE Custom Integrated Circuits Conference (CICC)},
  pages 1--4. IEEE, 2017.

\bibitem[Han et~al.(2015)Han, Mao, and Dally]{han2015deep}
Song Han, Huizi Mao, and William~J Dally.
\newblock Deep compression: Compressing deep neural networks with pruning,
  trained quantization and huffman coding.
\newblock \emph{arXiv preprint arXiv:1510.00149}, 2015.

\bibitem[He et~al.(2015)He, Zhang, Ren, and Sun]{he2015deep}
Kaiming He, XRSSJ Zhang, S~Ren, and J~Sun.
\newblock Deep residual learning for image recognition. eprint.
\newblock \emph{arXiv preprint arXiv:0706.1234}, 2015.

\bibitem[Hinton(2012)]{hinton2012neural}
G~Hinton.
\newblock Neural networks for machine learning. coursera,[video lectures],
  2012.

\bibitem[Hinton et~al.(2015)Hinton, Vinyals, and Dean]{hinton2015distilling}
Geoffrey Hinton, Oriol Vinyals, and Jeff Dean.
\newblock Distilling the knowledge in a neural network.
\newblock \emph{arXiv preprint arXiv:1503.02531}, 2015.

\bibitem[Howard et~al.(2017)Howard, Zhu, Chen, Kalenichenko, Wang, Weyand,
  Andreetto, and Adam]{howard2017mobilenets}
Andrew~G Howard, Menglong Zhu, Bo~Chen, Dmitry Kalenichenko, Weijun Wang,
  Tobias Weyand, Marco Andreetto, and Hartwig Adam.
\newblock Mobilenets: Efficient convolutional neural networks for mobile vision
  applications.
\newblock \emph{arXiv preprint arXiv:1704.04861}, 2017.

\bibitem[Hubara et~al.(2017)Hubara, Courbariaux, Soudry, El-Yaniv, and
  Bengio]{hubara2017quantized}
Itay Hubara, Matthieu Courbariaux, Daniel Soudry, Ran El-Yaniv, and Yoshua
  Bengio.
\newblock Quantized neural networks: Training neural networks with low
  precision weights and activations.
\newblock \emph{The Journal of Machine Learning Research}, 18\penalty0
  (1):\penalty0 6869--6898, 2017.

\bibitem[Ioffe and Szegedy(2015)]{ioffe2015batch}
Sergey Ioffe and Christian Szegedy.
\newblock Batch normalization: Accelerating deep network training by reducing
  internal covariate shift.
\newblock \emph{arXiv preprint arXiv:1502.03167}, 2015.

\bibitem[Jouppi et~al.(2017)Jouppi, Young, Patil, Patterson, Agrawal, Bajwa,
  Bates, Bhatia, Boden, Borchers, et~al.]{jouppi2017datacenter}
Norman~P Jouppi, Cliff Young, Nishant Patil, David Patterson, Gaurav Agrawal,
  Raminder Bajwa, Sarah Bates, Suresh Bhatia, Nan Boden, Al~Borchers, et~al.
\newblock In-datacenter performance analysis of a tensor processing unit.
\newblock In \emph{2017 ACM/IEEE 44th Annual International Symposium on
  Computer Architecture (ISCA)}, pages 1--12. IEEE, 2017.

\bibitem[Jung et~al.(2018)Jung, Son, Lee, Son, Kwak, Han, Hwang, and
  Choi]{jung2018learning}
Sangil Jung, Changyong Son, Seohyung Lee, Jinwoo Son, Youngjun Kwak, Jae-Joon
  Han, Sung~Ju Hwang, and Changkyu Choi.
\newblock Learning to quantize deep networks by optimizing quantization
  intervals with task loss, 2018.

\bibitem[LeCun et~al.(1990)LeCun, Denker, and Solla]{lecun1990optimal}
Yann LeCun, John~S Denker, and Sara~A Solla.
\newblock Optimal brain damage.
\newblock In \emph{Advances in neural information processing systems}, pages
  598--605, 1990.

\bibitem[Leroux et~al.(2019)Leroux, Vankeirsbilck, Verbelen, Simoens, and
  Dhoedt]{leroux2019training}
Sam Leroux, Bert Vankeirsbilck, Tim Verbelen, Pieter Simoens, and Bart Dhoedt.
\newblock Training binary neural networks with knowledge transfer.
\newblock \emph{Neurocomputing}, 2019.

\bibitem[Li et~al.(2016)Li, Zhang, and Liu]{li2016ternary}
Fengfu Li, Bo~Zhang, and Bin Liu.
\newblock Ternary weight networks.
\newblock \emph{arXiv preprint arXiv:1605.04711}, 2016.

\bibitem[Merolla et~al.(2016)Merolla, Appuswamy, Arthur, Esser, and
  Modha]{merolla2016deep}
Paul Merolla, Rathinakumar Appuswamy, John Arthur, Steve~K Esser, and
  Dharmendra Modha.
\newblock Deep neural networks are robust to weight binarization and other
  non-linear distortions.
\newblock \emph{arXiv preprint arXiv:1606.01981}, 2016.

\bibitem[Moons et~al.(2017)Moons, Uytterhoeven, Dehaene, and
  Verhelst]{moons201714}
Bert Moons, Roel Uytterhoeven, Wim Dehaene, and Marian Verhelst.
\newblock 14.5 envision: A 0.26-to-10tops/w subword-parallel
  dynamic-voltage-accuracy-frequency-scalable convolutional neural network
  processor in 28nm fdsoi.
\newblock In \emph{2017 IEEE International Solid-State Circuits Conference
  (ISSCC)}, pages 246--247. IEEE, 2017.

\bibitem[Polino et~al.(2018)Polino, Pascanu, and Alistarh]{polino2018model}
Antonio Polino, Razvan Pascanu, and Dan Alistarh.
\newblock Model compression via distillation and quantization.
\newblock \emph{arXiv preprint arXiv:1802.05668}, 2018.

\bibitem[Rastegari et~al.(2016)Rastegari, Ordonez, Redmon, and
  Farhadi]{rastegari2016xnor}
Mohammad Rastegari, Vicente Ordonez, Joseph Redmon, and Ali Farhadi.
\newblock Xnor-net: Imagenet classification using binary convolutional neural
  networks.
\newblock In \emph{European Conference on Computer Vision}, pages 525--542.
  Springer, 2016.

\bibitem[Redmon and Farhadi(2016)]{redmon2016yolo9000}
J~Redmon and A~Farhadi.
\newblock Yolo9000: Better, faster.
\newblock \emph{Stronger Preprint gr-qc/arXiv}, 1612, 2016.

\bibitem[Sainath and Parada(2015)]{sainath2015convolutional}
Tara Sainath and Carolina Parada.
\newblock Convolutional neural networks for small-footprint keyword spotting.
\newblock 2015.

\bibitem[Srivastava et~al.(2014)Srivastava, Hinton, Krizhevsky, Sutskever, and
  Salakhutdinov]{srivastava2014dropout}
Nitish Srivastava, Geoffrey Hinton, Alex Krizhevsky, Ilya Sutskever, and Ruslan
  Salakhutdinov.
\newblock Dropout: a simple way to prevent neural networks from overfitting.
\newblock \emph{The journal of machine learning research}, 15\penalty0
  (1):\penalty0 1929--1958, 2014.

\bibitem[Tang and Lin(2018)]{tang2018deep}
Raphael Tang and Jimmy Lin.
\newblock Deep residual learning for small-footprint keyword spotting.
\newblock In \emph{2018 IEEE International Conference on Acoustics, Speech and
  Signal Processing (ICASSP)}, pages 5484--5488. IEEE, 2018.

\bibitem[Warden(2017)]{warden2017speech}
Pete Warden.
\newblock Speech commands: A public dataset for single-word speech recognition.
\newblock \emph{Dataset available from http://download. tensorflow.
  org/data/speech\_commands\_v0}, 1, 2017.

\bibitem[Wu et~al.(2016)Wu, Wu, and Zhao]{wu2016binarized}
Xundong Wu, Yong Wu, and Yong Zhao.
\newblock Binarized neural networks on the imagenet classification task.
\newblock \emph{arXiv preprint arXiv:1604.03058}, 2016.

\bibitem[Xu et~al.(2018)Xu, Wang, Zhou, Lin, and Xiong]{xu2018deep}
Yuhui Xu, Yongzhuang Wang, Aojun Zhou, Weiyao Lin, and Hongkai Xiong.
\newblock Deep neural network compression with single and multiple level
  quantization.
\newblock In \emph{Thirty-Second AAAI Conference on Artificial Intelligence},
  2018.

\bibitem[Xue et~al.(2013)Xue, Li, and Gong]{xue2013restructuring}
Jian Xue, Jinyu Li, and Yifan Gong.
\newblock Restructuring of deep neural network acoustic models with singular
  value decomposition.
\newblock In \emph{Interspeech}, pages 2365--2369, 2013.

\bibitem[Zhang et~al.(2018)Zhang, Yang, Ye, and Hua]{zhang2018lq}
Dongqing Zhang, Jiaolong Yang, Dongqiangzi Ye, and Gang Hua.
\newblock Lq-nets: Learned quantization for highly accurate and compact deep
  neural networks.
\newblock In \emph{Proceedings of the European Conference on Computer Vision
  (ECCV)}, pages 365--382, 2018.

\bibitem[Zhang et~al.(2017)Zhang, Suda, Lai, and Chandra]{zhang2017hello}
Yundong Zhang, Naveen Suda, Liangzhen Lai, and Vikas Chandra.
\newblock Hello edge: Keyword spotting on microcontrollers.
\newblock \emph{arXiv preprint arXiv:1711.07128}, 2017.

\bibitem[Zhu et~al.(2016)Zhu, Han, Mao, and Dally]{zhu2016trained}
Chenzhuo Zhu, Song Han, Huizi Mao, and William~J Dally.
\newblock Trained ternary quantization.
\newblock \emph{arXiv preprint arXiv:1612.01064}, 2016.

\end{thebibliography}

\end{document}